\documentclass{article}
\usepackage[T1]{fontenc}
\usepackage[utf8]{inputenc}
\usepackage{float}
\usepackage{mathrsfs}
\usepackage{amsmath}
\usepackage{microtype}
\usepackage[unicode=true,
 bookmarks=false,
 breaklinks=false,pdfborder={0 0 1},backref=section,colorlinks=false]
 {hyperref}

\makeatletter

\newcommand{\lyxmathsym}[1]{\ifmmode\begingroup\def\b@ld{bold}
  \text{\ifx\math@version\b@ld\bfseries\fi#1}\endgroup\else#1\fi}

\floatstyle{ruled}
\newfloat{algorithm}{tbp}{loa}
\providecommand{\algorithmname}{Algorithm}
\floatname{algorithm}{\protect\algorithmname}

\usepackage{arxiv}

\usepackage{url}
\usepackage{booktabs}
\usepackage{amsfonts}
\usepackage{nicefrac}
\usepackage[noend]{algpseudocode}
\usepackage{algorithmicx}
\usepackage{algorithm}\usepackage{threeparttable}
\usepackage{cite}
\usepackage{url}

\title{A short note on the decision tree based \\ neural turing machine}
\author{%
Yingshi Chen 
}

\makeatother

\begin{document}
\title{An iterative K-FAC algorithm for Deep Learning}
\maketitle
\begin{abstract}
Kronecker-factored Approximate Curvature (K-FAC) method is a high
efficiency second order optimizer for the deep learning. Its training
time is less than SGD(or other first-order method) with same accuracy
in many large-scale problems. The key of K-FAC is to approximates
Fisher information matrix (FIM) as a block-diagonal matrix where each
block is an inverse of tiny Kronecker factors. In this short note,
we present CG-FAC --- an new iterative K-FAC algorithm. It uses conjugate
gradient method to approximate the nature gradient. This CG-FAC method
is matrix-free, that is, no need to generate the FIM matrix, also
no need to generate the Kronecker factors A and G. We prove that the
time and memory complexity of iterative CG-FAC is much less than that
of standard K-FAC algorithm. 
\end{abstract}

\section{Introduction}

Optimizer\cite{bottou2018optimization,goodfellow2016deep} is the
key component to reduce the training losses of deep learning\cite{goodfellow2016deep}.
Based on the the order of derivative, optimization methods are mainly
divided into two categories, first-order and second-order methods.
First-order method uses gradient information to update parameters(weights)
whereas second-order method would try to get fast convergence on the
addition curvature information. Based on the search space, optimization
methods can be divided into two categories: parameter space and distribution
space. Nature gradient method \cite{amari1998natural,amari2000adaptive,martens2020new}
is a second order method in the distribution space, which is proposed
by Shun-Ichi Amari in 1998 based on information geometry. Compare
to the most widely used first-order method (SGD,ADAM,...\cite{robbins1951stochastic,bottou2018optimization,kingma2014adam})
, Nature gradient contains more information in the distribution space,
so it would find a better solution with faster speed. In actual application,
it does face the same practical difficulties of second order methods.
It does need high-performance methods to get the inverse of hessian
matrix. The traditional matrix inversion algorithms(or other approximate
algorithms) require too much computation and memory resource tot be
implemented in practical applications. Until recent years, some novel
algorithms greatly reduce the memory usage and computational overhead.
Especially the Kronecker-factored Approximate Curvature (K-FAC) method
\cite{pauloski2020convolutional,grosse2016kronecker,martens2015optimizing}by
James Martens, etc. K-FAC is a method for efficiently approximating
the natural gradient in real deep learning problems. It approximates
the hessian matrix as Kronecker products of many smaller matrices,
which are more efficiently to deal with (for example, LU decomposition,
SVD decomposition,perform or other matrix operations) . 

In this short note, we present a new iterative K-FAC algorithm CG-FAC
on the special structure of hessian matrix in deep networks. This
method is matrix-free, that is, no need to generate the hessian matrix,
also no need to generate all small Kronecker factors. So it requires
much less calculation and memory usage than standard K-FAC algorithm.

\section{Background and notation \label{sec:background} }

In this section, we first give some concise definitions of optimization
problems of deep learning. Given training data $D={(x_{i},y_{i}),i=1,2,\cdots}$,
deep learning method tries to deduce the loss $\mathscr{\mathcal{L}}$
between prediction $\hat{y}=f(\theta:x)$ and the target $y$. At
each step of optimization process, the parameters \ensuremath{\theta}
would be updated along a direction $p$ and a step length $\eta$
\cite{nocedal2006numerical,absil2009optimization}:

\begin{equation}
\theta'=\theta+\eta p
\end{equation}

In the machine learning community, many methods have been proposed
to find the proper search direction $p$.

\subsubsection*{Stochastic gradient descent (SGD)}

A classical strategy is to choose $p$ along the steepest decent direction$\nabla_{\theta}\mathcal{L}$,
which is most widely used technology in the deep learning method.

\begin{equation}
p=\arg\min_{\left\Vert p\right\Vert \leq\epsilon}\mathit{\mathcal{L}}\left(\theta+p\right)=\nabla_{\theta}\mathcal{L}\label{eq:p}
\end{equation}

In the practical training process of deep learning, the data-sets
would always be randomly shuffled and spitted into many batches. The
optimization finds $p$ by back-propagation method in each mini-batch.
This stochastic gradient descent(SGD) method has may variants \cite{robbins1951stochastic,bottou2018optimization}
, including AdaGrad \cite{duchi2011adaptive}, RMSprop \cite{hinton2012neural},
and Adam \cite{kingma2014adam}.

Adagrad adapts the learning rate specifically to individual features:
that means that some of the weights in your data-set will have different
learning rates than others. This works really well for sparse data-sets
where a lot of input examples are missing. Adagrad has a major issue
though: the adaptive learning rate tends to get really small over
time. 

RMSprop is a special version of Adagrad developed by Professor Geoffrey
Hinton in his neural nets class. Instead of letting all of the gradients
accumulate for momentum, it only accumulates gradients in a fixed
window. RMSprop is similar to Adaprop, which is another optimizer
that seeks to solve some of the issues that Adagrad leaves open.

Adam stands for adaptive moment estimation, and is another way of
using past gradients to calculate current gradients. Adam also utilizes
the concept of momentum by adding fractions of previous gradients
to the current one. This optimizer has become pretty widespread, and
is practically accepted for use in training neural nets.

\subsubsection*{Quasi-Newton method}

Quasi-Newton methods \cite{nocedal2006numerical} use list of successive
gradients to approximate the inverse of Hessian matrix. \cite{bordes2009sgd,byrd2016stochastic,byrd2011use}
makes some attempts to combine the Broyden-Fletcher-GolfarbShanno
(BFGS) method and SGD method. Although Quasi-Newton method is widely
used in scientific computing, it is really difficult to apply in deep
learning. The number of parameters is very large. For example, AlexNet
has 60 million parameters and BERT\_large has 340 million parameters.
The standard Quasi-Newton methods need much more computation resource
than first-order method. And need much more time to converge.

\subsubsection*{Natural gradient method}

\cite{amari1998natural,amari2000adaptive,martens2020new} proposed
a new search direction (Natural gradient) in the distribution space.
That is, we are not only looking for suitable parameters, but also
for the distributions that reflect the essence of the problem more
deeply than parameters.The number and value of the parameters will
vary greatly, but the distribution should be always the same. To measure
the variance of distribution between steps, a common way is to use
Kullback--Leibler divergence\cite{kullback1997information} as a
metric of loss. As the loss gets smaller, the distribution changes
smaller and smaller. Finally, we not only get a solution with minimal
loss, but also a stable distribution corresponds to the problem. As
the difference between formula \ref{eq:p} and formula \ref{eq:KL}
shows: the natural gradient method replaces the Euclidean metric with
KL metric.

\begin{equation}
p=\arg\min_{\textrm{KL}\leq\epsilon}\mathit{\mathcal{L}}\left(\theta+p\right)\label{eq:KL}
\end{equation}

Let the hessian of KL metric is $G$. Then the second-order search
direction of formula\ref{eq:KL} is:
\begin{equation}
\widetilde{\nabla}_{\theta}\mathcal{L}=G^{-1}\nabla_{\theta}\mathcal{L}
\end{equation}

This is just the definition of natural gradient. \cite{watanabe2009algebraic}
pointed that the Fisher information matrix(FIM) is equal to the hessian
matrix of the Kullback--Leibler distance. So we would update the
parameters \ensuremath{\theta} by $F$

\begin{equation}
\theta'=\theta+\eta F^{-1}\nabla_{\theta}\mathcal{L}
\end{equation}

In the simplest case that $G$ is identity matrix, this is just the
standard steepest decent method shown in formula \ref{eq:p}.

In practical case of deep learning, the dimension of F is very large.
For example, AlexNet has 60 million parameters and BERT\_large has
340 million parameters. The standard second-order method would fail
for such huge matrices or would be very slow compared to first-order
method. A novel technique to get the inverse of F is the following
K-FAC method.

\subsubsection*{K-FAC method}

K-FAC is the shortcut for Kronecker-factored Approximate Curvature
(K-FAC). As the name suggests, this method approximates the FIM as
Kronecker products of smaller matrices, which are more efficiently
to deal with (LU decomposition, SVD decomposition...). Algorithm 1
gives the standard algorithm of this method. It would first generate
diagonal block matrix for each layer. The dimension of each block
is the number of parameters of corresponding layer. This decoupling
technique brings great convenience for the parallel and distribute
training. 

\begin{algorithm}
\caption{Standard Kronecker-factored Approximate of FIM}

\textbf{Input:} 

$\qquad$A deep neural network with $N$ layers

\hspace*{0.02in} \textbf{Output:} Approximate inverse of FIM

\begin{algorithmic}[1]

\State Call standard back propagation \cite{bishop1995neural} for
current batch. 

\State Get matrix $A_{i}$ and $G_{i}$ for $i^{th}$ layer on the
input $a_{i-1}$ and $g_{i}$(gradient of output):

\[
A_{i=}(a_{i-1}\cdot a_{i-1}^{T})\qquad G_{i=}(g_{i}\cdot g_{i}^{T})
\]

\State Block diagonalization: $N$ diagonal block for $N$ layers:

\[
F\cong\widehat{F}=\left[\begin{array}{cccc}
\hat{F}_{1}\\
 & \hat{F}_{2}\\
 &  & \ddots\\
 &  &  & \hat{F}_{N}
\end{array}\right]
\]

\State Approximates each diagonal block with the Kronecker product
of $A$ and $G$:

\[
\hat{F_{i}}=(a_{i-1}\cdot a_{i-1}^{T})\otimes(g_{i}g_{i}^{T})=A_{i-1}\otimes G_{i}
\]

\State Get the inverse of $\hat{F_{i}}$

\[
\hat{F_{i}}^{-1}=\left(A_{i-1}\otimes G_{i}\right)^{-1}=A_{i-1}^{-1}\otimes G_{i}^{-1}
\]

\end{algorithmic}
\end{algorithm}

\section{Iterative K-FAC algorithm \label{sec:sec_algorithm} }

At each mini-batch $\mathfrak{B}$, the classical SGD method would
update the current weight $\theta$ by

\begin{equation}
\mathbf{\theta'}=\mathbf{\theta}-\eta\frac{1}{\left|\mathfrak{B}\right|}\sum_{j\in\mathfrak{B}}\nabla_{\theta}\mathcal{L}_{j}
\end{equation}

where $\nabla_{\theta}\mathcal{L}_{j}$ is the gradient of loss function
$\mathscr{\mathcal{L}}$ at$j^{th}$ sample. 

To use the second-order curvature information and more information
from distribution space, we would update $\theta$ in the direction
of nature gradient, that is 

\begin{equation}
\mathbf{\theta'}=\mathbf{\theta}-\eta\frac{1}{\left|\mathfrak{B}\right|}F^{-1}\sum_{j\in\mathfrak{B}}\nabla_{\theta}\mathcal{L}_{j}
\end{equation}

As algorithm 1 shows: for the $i^{th}$ layer of network, let its
input is $a_{i-1}$(The activation of the$(i-1)^{th}$ layer), the
gradient of its output is $g_{i}$. K-FAC would update its weight
by $\hat{F}_{i}$(a diagonal block of F) \cite{amari1998natural,amari2000adaptive,martens2020new}

\begin{equation}
\hat{F_{i}}=(a_{i-1}\cdot a_{i-1}^{T})\otimes(g_{i}g_{i}^{T})=A_{i-1}\otimes G_{i}
\end{equation}
This is a Kronecker-factorization of $\hat{F_{i}}$ and $A_{i-1},G_{i}$
are the Kronecker factors.

In practical use, we usually use a dumped version to increase the
robustness. That is, add $\gamma I$to $\hat{F_{i}}$. Let $F_{\gamma}=\hat{F_{i}}+\gamma I,\nabla L=\frac{1}{\left|\mathfrak{B}\right|}\sum_{j\in\mathfrak{B}}\nabla_{\theta}\mathcal{L}_{j}$,
then the updating process is: 

\begin{equation}
\mathbf{\theta'}=\mathbf{\theta}-\eta F_{\gamma}^{-1}\nabla_{\theta}L
\end{equation}

So the main problem is to get the preconditioned gradient $F_{\gamma}^{-1}\nabla L$.
Let the unknown vector $x=F_{\gamma}^{-1}\nabla L$, then

\begin{equation}
F_{\gamma}x=\nabla L\label{eq:ax=00003Db}
\end{equation}

There are mainly two ways to solve equation \ref{eq:ax=00003Db}.
Let $n$ is the number of parameters in $i^{th}$ layer (that is,
$\nabla L\in R^{n}$). Let the dimension of $G_{i}$ is $n_{G}$ and
dimension of $A_{i-1}$ is $n_{A}$. We use these parameters to compare
the algorithm complexity of both methods. The first method is the
direct method, that is, to get some inverse form of $F_{\gamma}$.
For example, \cite{pauloski2020convolutional,grosse2016kronecker}
first get the eigendecompostion of$A_{i-1}$ and $G_{i}$, which need
$O(n_{A}^{3}+n_{G}^{3})$ operations. Then update the gradient, which
needs $O(2n(n_{A}+n_{G}))$. Therefore, the algorithm complexity is
$O(n_{A}^{3}+n_{G}^{3}+2n(n_{A}+n_{G}))$. And the memory complexity
is $O(n_{A}^{2}+n_{G}^{2}+n)$. The second method is the iterative
method. Or a finite step iteration in the Krylov space. Since $(\hat{F_{i}}+\gamma I)$
is positive definite (SPD) with proper $\gamma$, we would use standard
conjugate gradient (CG) method \cite{saad2003iterative,van1983matrix}
to solve this equation. So this method is called CG-FAC. The following
is the detail of this method: 

\begin{algorithm}[H]
\caption{CG-FAC method: approximate nature gradient with at most m iterations}
\hspace*{0.02in} \textbf{Input:} 

$\qquad n$: The parameters size in $i^{th}$ layer

$\qquad a_{i-1}$: The activation of the $(i-1)^{th}$ layer

$\qquad g_{i}$: The gradient of output in the $i^{th}$ layer

$\qquad\mathbf{b}=\nabla L$: The gradient of parameters in $i^{th}$
layer

$\qquad F_{\gamma}=\hat{F_{i}}+\gamma I$: The fisher information
matrix (FIM) with dumping parameter

$\qquad\mathbf{x}_{0}$: A guess of nature gradient (may use the value
from previous batch)

\begin{algorithmic}[1]

\Function{FV}{$v$}
\State $\theta$=$g_{i}^{T}va_{i-1}$
\State $v_{1}$=$g_{i}\theta a_{i-1}^{T}+\gamma v$
\State \Return $v_{1}$ 

\EndFunction

\subparagraph*{Conjugate gradient (CG) iteration to approximate nature gradient}

\State $\boldsymbol{p}_{0}=\boldsymbol{r}_{0}=\boldsymbol{b}-F_{\gamma}\boldsymbol{x}_{0}$

\State $\rho_{0}=\left\Vert \boldsymbol{r}_{0}\right\Vert ^{2}$

\For{$k=0,1,2,\cdots,m$}

\State \quad{}$\boldsymbol{u}_{k}=FV(\boldsymbol{p}_{k})$

\State\quad{}$s_{k}=\boldsymbol{p}_{k}\cdot\boldsymbol{u}_{k}$

\State \quad{}$\alpha_{k}=\rho_{k}/s_{k}$

\State \quad{}$\boldsymbol{x}_{k+1}=\boldsymbol{x}_{k}+\alpha\boldsymbol{p}_{k}$

\State \quad{}$\boldsymbol{r}_{k+1}=\boldsymbol{r}_{k}-\alpha\boldsymbol{u}_{k}$

\State \quad{}$\rho_{k+1}=\left\Vert \boldsymbol{r}_{k+1}\right\Vert ^{2}$

\State\quad{}If $\rho_{k+1}$is sufficiently small, then exit

\State \quad{}$\beta_{k}=\rho_{k+1}/\rho_{k}$

\State \quad{}$\boldsymbol{p}_{k+1}=\boldsymbol{r}_{k+1}+\beta_{k}\boldsymbol{p}_{k}$

\EndFor

\State \Return

$\boldsymbol{x}_{k+1}$: The approximation of nature gradient

\end{algorithmic}
\end{algorithm}

At each internal step, CG method needs one matrix-vector operations
($F_{\gamma}p_{k}$) and 5 vector operations (in the sense of BLAS
vector operation). As the function FV$(v)$ shows, there are some
fast methods on the special structure of $A_{i-1}$ and $G_{i}$.
Each FV needs $O(3n)$ operations. So the algorithm complexity of
each step is $O(8n)$. The CG needs at most n steps to converge \cite{van1983matrix,saad2003iterative}.
In the real application of CG method, only need $m\ll n$ steps \cite{van1983matrix,saad2003iterative}.
For example, in the training process from batch to batch, the difference
between gradients is small. We could use the nature gradient of previous
batch as the initial guess of CG method. Then only needs 10-20 internal
steps. Therefore, CG-FAC needs $O(m(8n))$ operations, which is much
less than$O(n_{A}^{3}+n_{G}^{3}+2n(n_{A}+n_{G}))$ operations in the
direct methods. As the function FV shows, there is no need to generate
the matrix $F_{\gamma}$, $A_{i-1}$ and $G_{i}$ in the iterative
method. This method is matrix-free. Therefore, the memory complexity
is CG-FAC is only $O(n_{A}+n_{G}+n)$, which is also less than the
$O(n_{A}^{2}+n_{G}^{2}+n)$ of direct method. We would expect better
data locality and higher cache hit ratio in the iterative method. 

\section{Conclusion}

In this short note, we present an iterative K-FAC algorithm on the
special structure of Fisher information matrix (FIM). This method
is matrix-free, that is, no need to generate FIM, also no need to
generate the Kronecker factors A and G. We will give more detailed
experiments in later papers.

\bibliographystyle{unsrt}

\end{document}